\documentclass{article} 
\usepackage{iclr2021_conference,times}

\usepackage{amsmath,amsfonts,bm}

\def\eqref#1{equation~\ref{#1}}

\def\1{\bm{1}}

\DeclareMathAlphabet{\mathsfit}{\encodingdefault}{\sfdefault}{m}{sl}
\SetMathAlphabet{\mathsfit}{bold}{\encodingdefault}{\sfdefault}{bx}{n}

\usepackage{hyperref}
\usepackage{url}
\usepackage{graphicx}
\usepackage[ruled,linesnumbered]{algorithm2e}
\usepackage{amssymb}
\usepackage{multirow}
\usepackage{booktabs}
\usepackage{wrapfig}

\title{Unbiased IoU for \\ Spherical Image Object Detection}

\iclrfinalcopy

\author{Qiang Zhao$^1$, \quad Bin Chen$^1$, \quad Hang Xu$^2$, \quad Yike Ma$^1$ \\
\AND
Xiaodong Li$^3$, \quad Bailan Feng$^3$, \quad Chenggang Yan$^2$, \quad Feng Dai$^{1*}$ \\
\\
$^1$Institute of Computing Technology, Chinese Academy of Sciences, Beijing, China\\
$^2$Hangzhou Dianzi University, Hangzhou, China\\
$^3$Huawei Technologies Co., Ltd\\
\texttt{\{zhaoqiang,chenbin20s,ykma,fdai\}@ict.ac.cn} \\
\texttt{\{hxu,cgyan\}@hdu.edu.cn} \\
\texttt{\{lixiaodong33,fengbailan\}@huawei.com}
}

\begin{document}

\maketitle

\begin{abstract}
As one of the most fundamental and challenging problems in computer
vision, object detection tries to locate object instances and
find their categories in natural images. The most important step
in the evaluation of object detection algorithm is calculating the
intersection-over-union (IoU) between the predicted bounding box
and the ground truth one. Although this procedure is well-defined
and solved for planar images, it is not easy for spherical image
object detection. Existing methods either compute the IoUs based
on biased bounding box representations or make excessive approximations,
thus would give incorrect results. In this paper, we first
identify that spherical rectangles are unbiased bounding boxes for
objects in spherical images, and then propose an analytical method
for IoU calculation without any approximations. Based on the unbiased
representation and calculation, we also present an anchor free
object detection algorithm for spherical images. The experiments
on two spherical object detection datasets show that the proposed
method can achieve better performance than existing methods.
\end{abstract}

\section{Introduction}

Due to the development of image stitching techniques \cite{stitching}, numerous economic panoramic cameras have been developed in the recent years, such as Ricoh Theta, Samsumg Gear360, Insta 360, etc. With just a few clicks, these cameras can rapidly and easily capture panoramic (spherical) images or videos, which have $360^\circ$ field of view and provide a strong sense of presence. Thus these new type of multi media data are widely used in virtual navigation, cultural heritage and entertainment industry \cite{StreetView}. As one of the most fundamental and challenging problems in computer vision, object detection tries to determine the category, location and extent of each object instance appearing in an image. With the growing amount of panoramic data, it is also required to detect objects in spherical images for better understanding their content. For example, Hu et al. treat detected foreground objects as the targets to be followed in $360^\circ$ piloting \cite{Pilot}. 

\begin{wrapfigure}{r}{0.68\textwidth}
  \begin{center}
    \includegraphics[width=\linewidth]{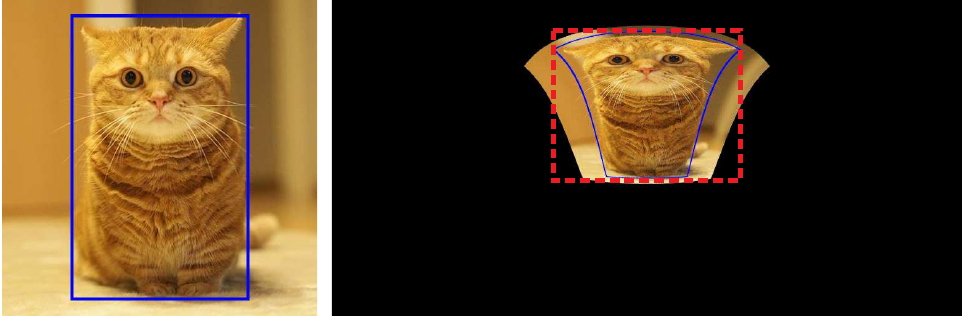}
	\caption{Axis-aligned rectangle used in planar image object detection (left) can not tightly bound object in spherical image (right).}
	\label{fig:difference}
  \end{center}
\end{wrapfigure}

In the literature, a lot of planar image object detection algorithms \cite{FasterRCNN,CenterNet} have been proposed and this field has achieved significant breakthroughs thanks to the emergence of deep learning techniques. However, comparatively limited studies exist for object detection fo spherical images. This is because equirectangular projection\footnote{It is the mostly used format by camera vendors and panoramic data sharing web sites.} would introduce image distortions. These image distortions have made two challenges to spherical image object detection tasks. First, the traditional convolutional layers are not applicable as they assume that the underlying image data have a regular planar grid structure. Second, the evaluation criteria adopted in planar object detection is not suitable because axis-aligned rectangles can not tightly bound objects in spherical images as shown in Figure \ref{fig:difference}. All these two problems come from the fact that the spherical images are signals defined on the sphere. Although the first problem has been solved to some degree \cite{SphFeature,DistortionAware,SphereNet,SpherePHD,TangentImage}, the latter one is unsolved. Some works directly use axis-aligned rectangles as bounding boxes for objects in spherical images \cite{DetectProject,DetectMultikernel} and check whether a predicted detection is correct based on the intersection over union (IoU) between two rectangles. This simple strategy would give large errors especially for objects near the polar regions that have serious distortions. Other works utilize the rectangles on tangent planes of sphere as bounding boxes \cite{SphFeature,SphereNet}. Although this strategy does not suffer from image distortions, it brings additional challenge for IoU calculation: the two rectangles involved in IoU calculation may fall on different tangent planes, which makes the traditional IoU calculation unapplicable. Lee et al. exploited circles on the spherical image as bounding boxes, which would also give biased result \cite{SpherePHD}. The works in \cite{360Indoor} and \cite{AAAICriteria} represent each object in spherical images using a spherical rectangle, which conforms to the imaging sphere model. However, when computing the IoU, they either convert the spherical rectangles to rectangles \cite{360Indoor} or consider the spherical rectangles as part of spherical zones \cite{AAAICriteria}, which would give incorrect results. In summary, none of existing methods give \emph{unbiased} representation and IoU calculation for spherical image object detection task.


In this paper, we first illustrate that the spherical rectangles are unbiased representation for objects in spherical images by making the analogy from planar case. Then we give the unbiased IoU calculation using spherical geometry. Significantly, our unbiased IoU is not only be used to derive the APs for performance evaluation, but also plays an important role in ground truth generation discussed in Section \ref{sec:impl_det}. Please note that our calculation does not make any \textit{approximations}. Our method also does not compute the area of intersection and union by integral on sphere, as the accuracy of this strategy is dependent on the resolution of the underlying spherical image. Instead, we give analytical solutions in this paper. Based on the new representation and calculation, we propose an anchor-free object detection algorithm for spherical images. Our method simply resembles the idea of CenterNet \cite{CenterNet}, but explicitly considers the geometry for spherical images. Specifically, we revisit the ground truth generation and loss function design for spherical case. We also replace the traditional convolutional layers with distortion aware spherical convolutional layers. For evaluation, we carry out experiments on three spherical datasets, including one real-world dataset and two synthetic datasets. It shows that our method can get better performance than baseline methods.

The rest of this paper is organized as follows: Section \ref{sec:related} reviews the most related work. We introduce the unbiased IoU calculation in Section \ref{sec:iou}, which is followed by the details of our proposed spherical object detection method in Section \ref{sec:method}. Section \ref{sec:experiment} gives the experimental results. Section \ref{sec:conclusion} concludes the paper and gives the possible directions of future work. 

\section{Related Work}\label{sec:related}
In this section, we discuss some works related to this paper, including planar image object detection, spherical image object detection and spherical object detection dataset.

\subsection{Planer Image Objection Detection}
There are numerous object detection algorithms for planar images. Here we only briefly introduce some representative ones. The readers are referred to \cite{Survey} for a good survey. Faster R-CNN framework \cite{FasterRCNN} is composed of two modules, where the first module is a deep fully convolutional network that proposes regions, and the second module is the Fast R-CNN detector \cite{FastRCNN} for region classification and bounding box regression. YOLO \cite{YOLO} frames object detection as a regression problem to spatially separated bounding boxes and associated class probabilities. The bounding boxes and class probabilities can be predicted directly from full images in one evaluation. Therefore, it can be optimized end-to-end directly on detection performance. SSD \cite{SSD} predicts category scores and box offsets for a fixed set of default bounding boxes using small convolutional filters applied to multi-scale feature maps. Thus it is faster than YOLO and with an accuracy competitive with Faster R-CNN. By formulating bounding box object detection as detecting paired top-left and bottom-right keypoints, CornerNet \cite{CornerNet} eliminates the need for designing a set of anchor boxes commonly used in prior detectors. CenterNet \cite{CenterNet} goes further and models an object as a single point, i.e. the center point of its bounding box. Then, it uses keypoint estimation to find center point and regresses to all other object properties, such as size and location.

\subsection{Spherical Image Objection Detection}
The object detection for spherical images is challenging due to the image distortions. 
\cite{DetectProject} transforms spherical image into four sub-images through stereographic projection, then YOLO is applied on each sub-image for object detection. The detected objects on each sub-image are back-projected to spherical image. However stereographic projection would also bring image distortions. 
\cite{DetectMultikernel} uses Faster R-CNN to detect objects in spherical images. To alleviate the image distortions, it applies a multi-kernel layer after ROI Pooling layer and adds the position information of each object region proposal into the network. \cite{SphFeature} projects the feature maps extracted by spherical convolutional layers to the tangent plane, and applies the planar proposal network on the tangent plane for object detection. \cite{SphereNet} proposes the spherical single shot multi-box detector, which adapts the popular SSD \cite{SSD} to spherical images. Anchor boxes are now placed on tangent planes of the sphere and are defined in terms of spherical coordinates of point of tangency, the width and height of the box on the tangent plane. \cite{SpherePHD} performs the vehicle detection based on YOLO \cite{YOLO} architecture with CNNs on spherical polyhedron representation of panoramic images. \cite{360Indoor} presents a real-world $360^\circ$ panoramic object detection dataset, and conducts widely used planar object detection approaches on this dataset for performance evaluation. \cite{AAAICriteria} proposes reprojection R-CNN, which is a two-stage $360^\circ$ object detector. It contains two stages, where the first stage is a spherical region proposal network that efficiently proposes coarse detections on spherical images, and the second stage is a reprojection network that accurately refines the proposals by applying another planar detector on the rectified spherical proposal.

\subsection{Spherical Object Detection Dataset}
There are several types of methods to prepare dataset for spherical object detection. The first type of methods transform the planar datasets and annotations to panoramic ones \cite{SphFeature,AAAICriteria}. The second type of methods composite real world background spherical images with rendered images \cite{SphereNet} or segmented images \cite{AAAICriteria}. As the composition configuration are set by ourselves, the bounding box parameters of composited objects can be easily obtained. Another approach generates realistic synthetic images with pixel-level annotations from virtual world. For example, the SYNTHIA dataset \cite{SYNTHIA}, a vitrual road driving image dataset, gives the ground truth labels for each object in each scene. The last type of methods capture spherical images and manually annotate the objects. The OmPaCa dataset \cite{SphereNet} is a real world dataset of omnidirectional images of real cars. The 360GoogleStreetView dataset \cite{DetectProject} contains images from Google Street View, and has labeled distorted people and cars. The WHU dataset \cite{Sensors} consists of images captured from vehicle-borne Ladybug camera, and four classes of objects of interest are manually labeled. The 360-Indoor dataset \cite{360Indoor}, which consists of complex indoor images containing common objects, is a new benchmark for visual object detection and class recognition in $360^\circ$ spherical indoor images.

\section{Spherical IoU}\label{sec:iou}
In this section, we first illustrate that the spherical rectangles are natural representation for objects in spherical images, then discuss why existing criteria are biased. Finally, we introduce our unbiased IoU calculation for spherical image object detection.

\begin{figure}[!t]
	\centering
	\includegraphics[width=0.9\linewidth]{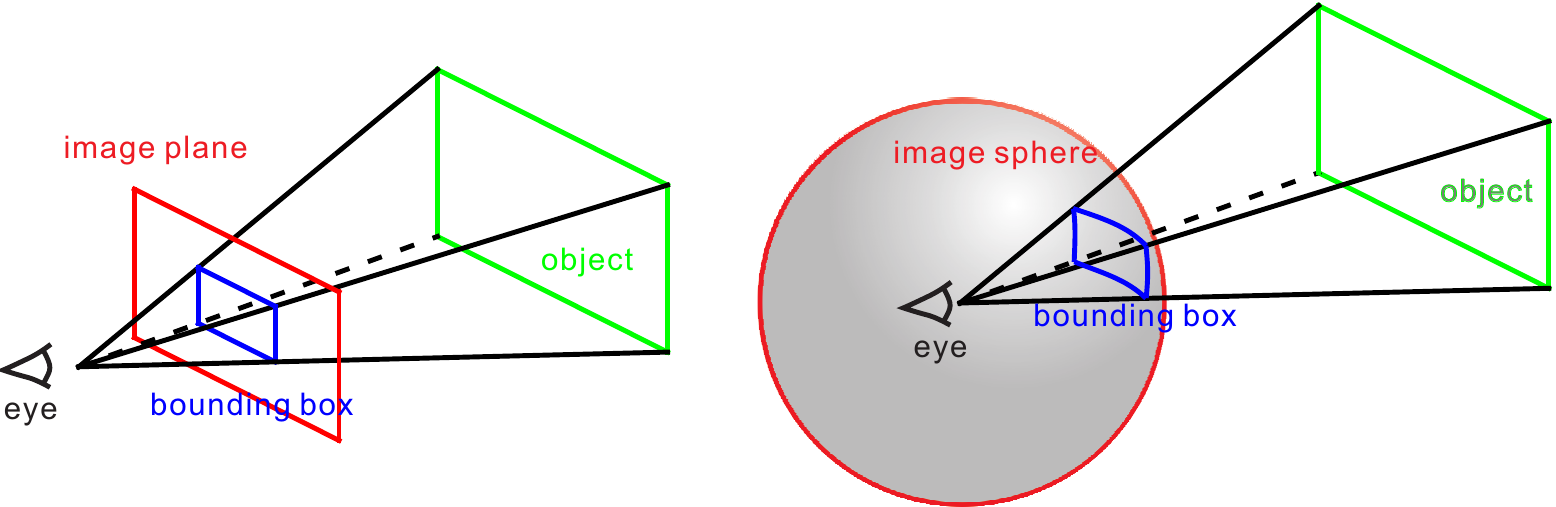}
	\caption{The bounding box of objects in planar image and spherical image. Please see text for detailed discussion.}
	\label{fig:camera}
\end{figure}

\subsection{Bounding Box Representation}
For the evaluation of generic object detection algorithms, one of the most important problem is how to represent the objects in images. In planar case, the spatial location and extent of an object are usually defined coarsely using an axis-aligned rectangle $(x,y,w,h)$ tightly bounding the object, where $(x,y)$ is the center point of the rectangle and $(w,h)$ is the width and height of the rectangle respectively. The rectangle is formed by the intersection between the image plane and the four surrounding faces of the viewing frustum that contains the 3D object as shown in Figue \ref{fig:camera}. By making an analogy, we can think that the bounding box for object on spherical image is formed by the intersection between the image sphere and the viewing frustum. As the four planes corresponding to the faces of the viewing frustum all pass through the center point of the sphere, thus the bounding box is a spherical rectangle. In this paper, we use $(\theta,\phi,\alpha,\beta)$ to denote a spherical rectangle, where $\theta$ is the azimuthal angle, $\phi$ is the polar angle, $\alpha$ and $\beta$ is the horizontal and vertical field of view respectively. Although we can also use the lengths of the great-circle arcs to represent the extent of the spherical rectangle, it is not convenient for the following IoU calculation.

\subsection{Existing Evaluation Criteria}\label{sec:ExistingCriteria}
In this section, we compare our representation with that adopted by previous works and show that all existing representations are biased. 

\begin{figure}[!t]
	\centering
	\includegraphics[width=0.99\linewidth]{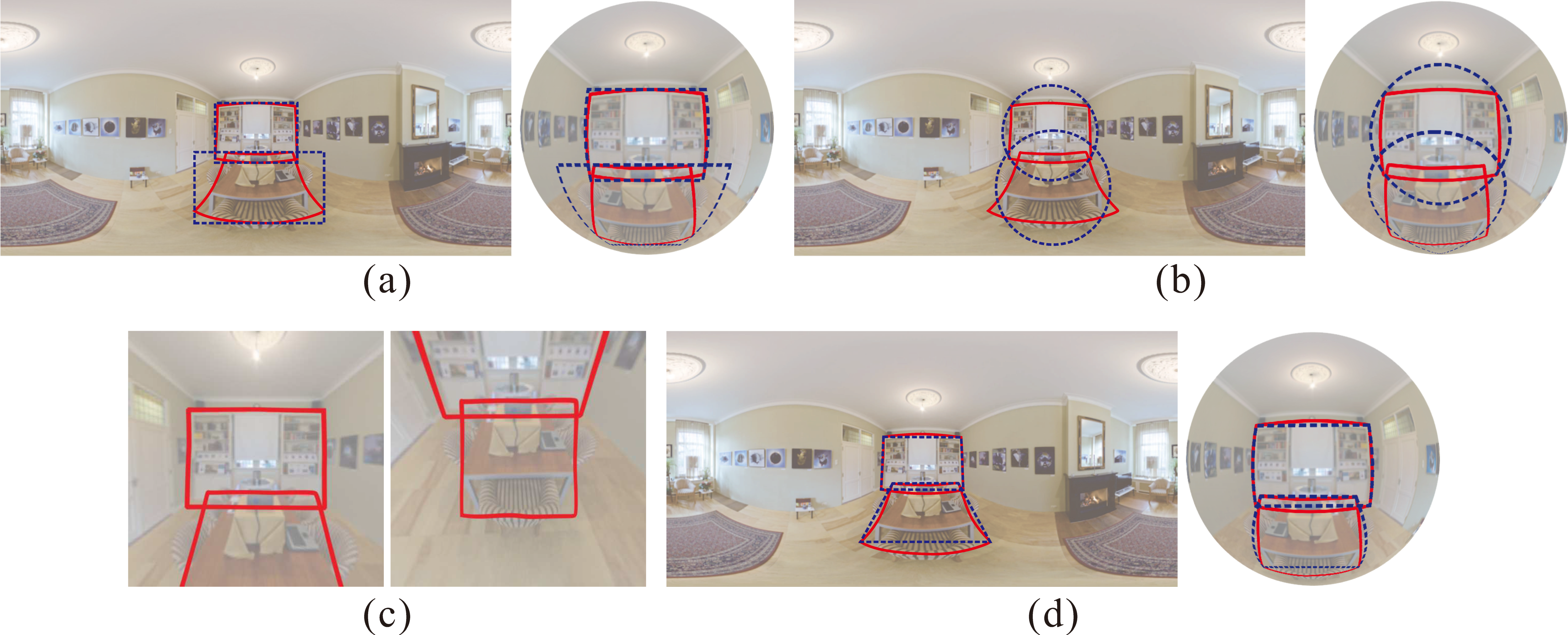}
	\caption{Existing representations and evaluation criteria for spherical image object detection: (a) using axis-aligned rectangles on spherical image, (b) using circles on spherical image, (c) using axis-aligned rectangles on tangent planes, (d) using axis-aligned rectangles on tangent planes but computing IoU based on the intersection of two polygons on spherical image. In each sub-figure, the red curves are the spherical rectangles used in this paper.}
	\label{fig:BBComparison}
\end{figure}

The works in \cite{DetectProject} and \cite{DetectMultikernel} use axis-aligned rectangles to represent objects in spherical images as shown in Figure \ref{fig:BBComparison} (a). Then they compute IoU based on intersection between two rectangles. However, there would be large errors. Circles are used to represent spherical objects in \cite{SpherePHD}, where each circle is denoted by its center and radius. IoU is computed based on intersection between two circles. This method also has large errors as shown in Figure \ref{fig:BBComparison} (b).

A plausible way to represent objects in spherical images is using axis-aligned rectangles on the tangent plane \cite{SphFeature,SphereNet}. However, it is challenging to compute IoU based on this representation, as it is unlikely that the estimated bounding box and the ground truth fall on the same tangent plane except they have the same center point. For example, we can project the spherical image to the tangent plane with the center point of one object as the point of tangency. As shown in Figure \ref{fig:BBComparison} (c), the bounding box of this object would be rectangle on the tangent plane, while the bounding boxes of other objects are not rectangles. Therefore, we can not compute the IoU based on the intersection between two rectangles. To deal with this problem, \cite{SphereNet} samples evenly spaced points along the rectangular bounding box on the tangent plane and projects them to spherical image. Then IoU can be computed based on the intersection of two constructed polygonal regions as shown in Figure \ref{fig:BBComparison} (d). However it is just an approximation to the unbiased solution and its accuracy is highly dependent on the point sampling density. Furthermore, the intersection is computed on the spherical image and suffers from distortions.  

The works in \cite{360Indoor} and \cite{AAAICriteria} also use unbiased representation, i.e. spherical rectangle, as our work. However, they do not give unbiased correct IoU calculation. As \cite{360Indoor} just uses planar object detector on their collected real-world $360^\circ$ object detection dataset, they directly applies planar rectangle based IoU calculation for performance evaluation. Although \cite{AAAICriteria} realize that the IoU should be calculated directly on sphere instead of on spherical image, their solution has made too many approximations and thus is also biased. The first approximation is that they treat spherical rectangles as parts of spherical zones\footnote{They term the shape as spherical segment, but the surface of the spherical segment consists of spherical zone and two bases.}. Actually, the area they calculated is the spherical area of the rectangle on the spherical image, i.e. the area of the blue shaper on the right image of Figure \ref{fig:BBComparison} (a). We can get this from the fact that circles of latitude correspond to a set of parallel planes, and spherical zone is defined by cutting a sphere with a pair of parallel planes. The second approximation they made is that they assume that the intersection between two spherical rectangles also forms a spherical rectangle. This assumption is excessive and incorrect. In the following, we will give our unbiased IoU calculation without approximations.

\subsection{Unbiased IoU Calculation}
Given two bounding boxes $b_1$ and $b_2$ represented by spherical rectangles $(\theta_1,\phi_1,\alpha_1,\beta_1)$ and $(\theta_2,\phi_2,\alpha_2,\beta_2)$, their IoU can be computed as
\begin{equation}
IoU(b_1,b_2)=\frac{A(b_1\cap b_2)}{A(b_1\cup b_2)}=\frac{A(b_1\cap b_2)}{A(b_1)+A(b_2)-A(b_1\cap b_2)},
\end{equation}
where $A(\cdot)$ is the area of the shape. Therefore, the IoU calculation can be formulated as the problem that computes the area of each spherical rectangle and the intersection between two spherical rectangles. One direct method for area calculation is using integral. For example, we can leverage the equation given in \cite{Segmentation} to compute the surface areas for pixels of the spherical image, and take their sum to get the values for $A(b_1)$, $A(b_2)$ and $A(b_1\cap b_2)$. However, the accuracy is dependent on the resolution of spherical image, especially for the pixels falling on the boundaries of spherical rectangles. Here, we seek to find the analytical solution for unbiased IoU calculation. The calculation of the area of each spherical rectangle is relatively easy and it can be obtained by
\begin{equation}\label{equ:area}
A(b_i)=4\arccos(-\sin\frac{\alpha_i}{2}\sin\frac{\beta_i}{2})-2\pi, \text{for}\ i\in\{1,2\}.
\end{equation}
The derivation is given in the supplementary material.

\begin{figure}[!t]
	\centering
	\includegraphics[width=0.99\linewidth]{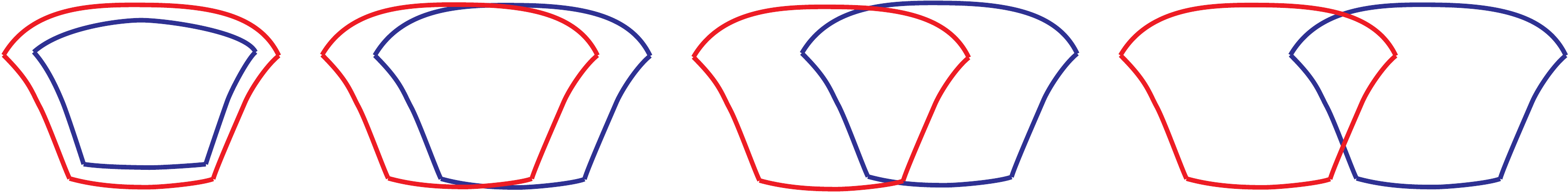}
	\caption{The intersection between two spherical rectangles may have different shapes. From left to right, the intersection is spherical rectangle, 6-sided spherical polygon, 5-sided spherical polygon and 4-sided spherical polygon respectively. Please note that here we only give some example cases and there exist intersections with other shapes.}
	\label{fig:intersection}
\end{figure}

The calculation of the area of intersection $A(b_1\cap b_2)$ between two spherical rectangles is very complex. This is because the intersection region may be not a spherical quadrangle, not to mention that it is not a spherical rectangle. We show the complexity in Figure \ref{fig:intersection}. As the boundaries of the intersection $b_1\cap b_2$ are all  great-circle arcs, we can use the following formula to computer the area of intersection
\begin{equation}\label{equ:polygon}
A(b_1\cap b_2)=\sum_{i=1}^{n}\omega_i-(n-2)\pi,
\end{equation}
if the intersection is $n$-sided spherical polygon. In the equation, $\omega_i$ is the angle of the spherical polygon, which equals to the angle between the planes that adjacent boundaries fall on. Then the core problem becomes determining the number $n$ of boundaries and finding which spherical rectangle each boundary comes from. Although the algorithm proposed in \cite{ComputationalGeometry} may be used to solve the problem, here we introduce a simpler and more robust one. 

\begin{algorithm}
	\caption{Intersection Area Computation}\label{algorithm}
	\KwIn{Two spherical rectangles $b_1$ and $b_2$ denoted as $(\theta_1,\phi_1,\alpha_1,\beta_1)$ and $(\theta_2,\phi_2,\alpha_2,\beta_2)$ }
	\KwOut{the area of intersection $A(b_1\cap b_2)$}
	\If{$b_1\cap b_2=\emptyset$}{\Return 0\;}
	\If{$b_1\subset b_2\ \mathrm{or}\ b_2\subset b_1$}{\Return $\min(A(b_1),A(b_2))$\;}
	compute the vertices $\mathcal{V}_i$ of spherical rectangle $b_i$\;
	compute the set $\mathcal{P}$ of intersection points between boundaries of $b_1$ and those of $b_2$\;
	$\mathcal{P}\leftarrow \mathcal{P}\cup\mathcal{V}_1\cup\mathcal{V}_2$\;
	remove the points $p$ in $\mathcal{P}$ such that $p\notin b_1$ or $p\notin b_2$\;
	remove duplicated points in $\mathcal{P}$ via loop detection\;
	\For{$p_i\in\mathcal{P}$}{compute the angle $\omega_i$}
	\Return $A(b_1\cap b_2)$ computed via Equation \ref{equ:polygon}\;
\end{algorithm}

Our method first checks whether there are no intersection between two spherical rectangles or whether one spherical rectangle is inside of the other. If so, $A(b_1\cap b_2)=0$ or $A(b_1\cap b_2)=\min(A(b_1),A(b_2))$. Otherwise, we compute all the intersection points between the boundaries of spherical rectangle $b_1$ and those of $b_2$. We remove the intersection points that fall outside of $b_1$ or $b_2$, then the area of intersection between $b_1$ and $b_2$ can be computed via Equation \ref{equ:polygon}. The other thing we should consider is that more than two boundaries may intersect at the same point, this case can be easily dealt with via loop detection and finally we can get the intersection area. The outline of our method is given in Algorithm \ref{algorithm}.

\section{Our Method}\label{sec:method}
\begin{figure}[!t]
	\centering
	\includegraphics[width=0.99\linewidth]{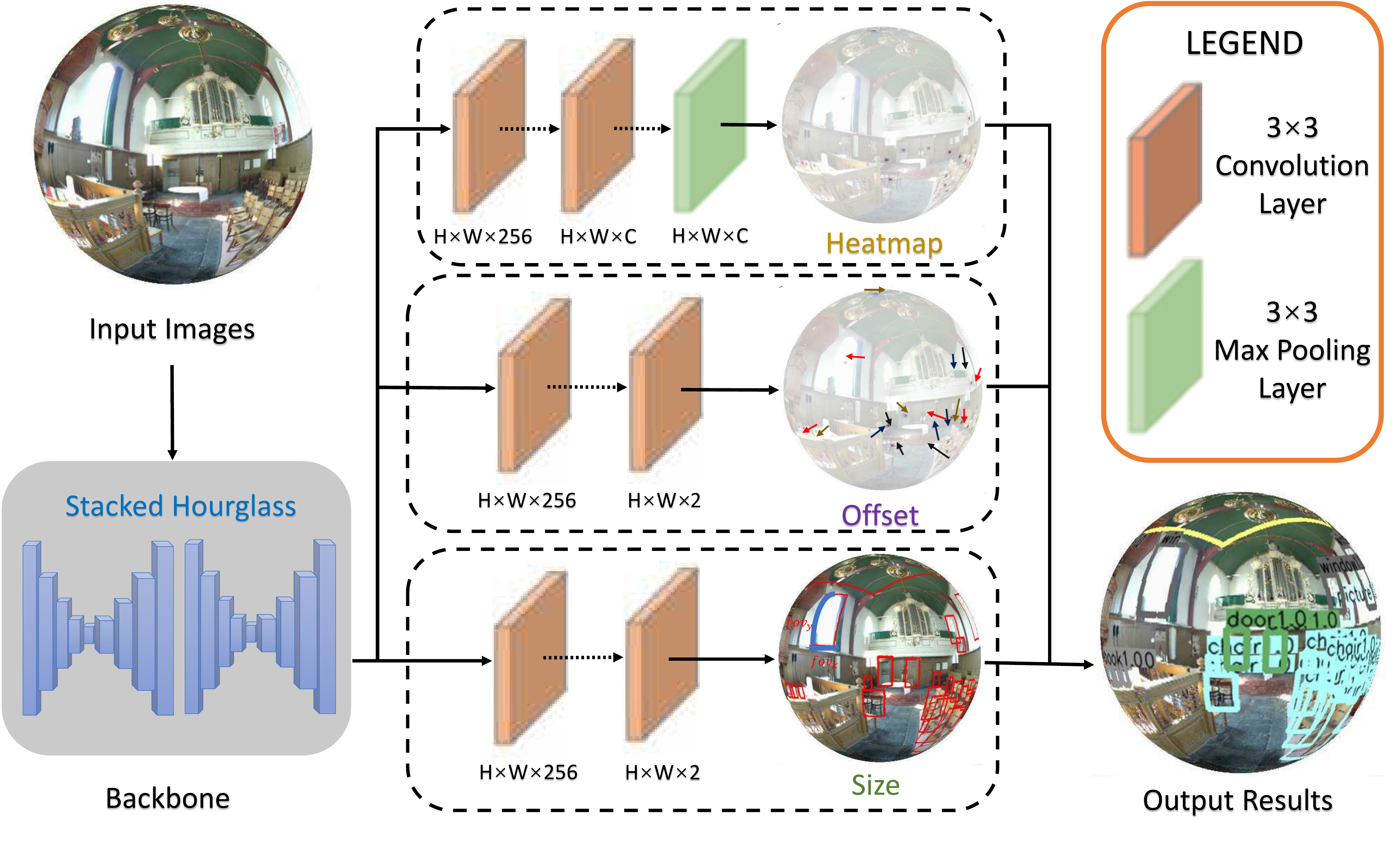}
	\caption{Our network takes spherical images as input and predicts heatmaps, offsets and sizes. With these information, we can determine the spherical rectangle bounding box for each object in the input spherical image.}
	\label{fig:network}
\end{figure}

Because the calculation of the area of spherical rectangles is more complex, it is not easy to define reference spherical rectangles with different scales and aspect ratios for anchor-based object detection algorithms \cite{FasterRCNN}. Therefore, here we propose an anchor-free object detector for spherical images based on CenterNet \cite{CenterNet} and make it applicable for spherical object detection.

\subsection{Spherical CenterNet}\label{sec:centernet}
Given a spherical image $I$ in Equirectangular Projection format, our goal is to predict the center point $(\theta_i,\phi_i)$ and the field of view $(\alpha_i,\beta_i)$ of bounding box for each object $i\in\left\lbrace1,2,\cdots,N \right\rbrace $ in image $I$. This is accomplished by using a convolutional network called Spherical CenterNet shown in Figure \ref{fig:network}. Note that in order to reflect the object detection task on spherical images, we use original spherical images as input images and output results in Figure \ref{fig:network}, instead of spherical images in Equirectangular Projection format, which is the actual inputs of the Spherical CenterNet network.

The input spherical image is first processed by a backbone network, whose output is fed into three branches for spherical bounding boxes prediction. The first branch produces a heatmap $p\in[0,1]^{W\times H\times C}$ for center points of all objects, where $W\times H$ is the size of the heatmap and $C$ is the number of object categories. The score $p_{xyc}$ at location $(x,y)$ for class $c$ indicates the possibility that the point $(x,y)$ is the center point of a spherical object belonging to category $c$. The input spherical image $I$ would be downsampled by the backbone for global information extraction, e.g. 4 times smaller when using stacked hourglass network. This will lead to discretization errors when we remap the locations from the heatmap $p$ to the input image $I$. To address this problem, the second branch predicts local offset $\mathbf{o}_i=(\Delta\theta_i,\Delta\phi_i)$ to slightly adjust the location of center point of each object $i$. The last branch is used to estimate the field of view $\mathbf{s}_i=(\alpha_i,\beta_i)$ of the spherical bounding box. 

Based on the architecture of the above three branches, we design our overall training objective as 
\begin{equation}
L=L_{cls}+\lambda_{off}L_{off}+\lambda_{fov}L_{fov},
\end{equation}
where $L_{cls}$ is the classification loss, $L_{off}$ and $L_{fov}$ are the regression loss for offset and field of view respectively. $\lambda_{off}$ and $\lambda_{fov}$ are the weights for the last two terms. The loss $L_{cls}$ is similar to that of CornerNet \cite{CornerNet} and CenterNet \cite{CenterNet}, and is based on focal loss \cite{FocalLoss}
\begin{equation}\label{equ:CELoss}
L_{cls}=\frac{-1}{N}\sum_{xyc}w_{xy}\begin{cases}
(1-p_{xyc})^2\log(p_{xyc})&\text{if}\ y_{xyc}=1,\\
(1-y_{xyc})^4(p_{xyc})^2\log(1-p_{xyc})&\text{otherwise},
\end{cases}
\end{equation}
where $y_{xyc}$ is the value of ground truth heatmap, whose generation will be described in the following section. A little difference is that we introduce a weight for each pixel at location $(x,y)$. The pixels near the polar region, which are more distorted, have smaller weights than the pixels near the equatorial region, which are less distorted. The weights $w_{xy}$ are computed based on the surface area of the pixels on unit sphere
\begin{equation}\label{equ:weight}
w_{xy}=\left( \cos\frac{y\pi}{H}-\cos\frac{(y+1)\pi}{H}\right) \frac{2\pi}{W}.
\end{equation}
As the center points of objects all fall on the curved spherical surface, we measure the offset loss with the angle between two 3D unit vectors
\begin{equation}\label{equ:offLoss}
L_{off}=\frac{1}{N}\sum_{i}\arccos\left(\langle\mathcal{T}\left(\mathbf{c}_i+\mathbf{o}_i \right),\mathcal{T}\left(\mathbf{c}_i+\mathbf{\hat{o}}_i \right)\rangle  \right), 
\end{equation}
where $\mathbf{c}_i=(\theta_i,\phi_i)$ is the center point of object $i$, $\mathbf{\hat{o}}_i$ is the ground truth offset, $\mathcal{T}(\cdot)$ is the transformation that convert the azimuthal and polar angle to 3D unit vector in Cartesian coordinate system, $\langle\cdot,\cdot\rangle$ is the dot product of two input vectors. For field of view regression, we simply use the L1 loss
\begin{equation}\label{equ:fovLoss}
L_{fov}=\frac{1}{N}\sum_{i}\left|\mathbf{s}_i- \mathbf{\hat{s}}_i\right|,
\end{equation}
where $\mathbf{\hat{s}}_i=(\hat{\alpha}_i,\hat{\beta}_i)$ is the ground truth field of view for object $i$. Please note that we do not incorporate the weights $w_xy$ in the design of $L_{off}$ and $L_{fov}$. This is because the supervisions of these two terms only act at center point locations, while the loss $L_{cls}$ takes sum over all locations.

During inference, we first extract the peaks in the predicted hearmap and only keep the top $100$ peaks, whose value is greater than its neighbors. Let $(x_i,y_i)$ be the coordinates of the peaks. We then transform them to azimuthal and polar angles according to the definition of Equirectangular projection, i.e. $\theta_i=\frac{2x_i\pi}{W}, \phi_i=\frac{y_i\pi}{H}$. Finally, the bounding boxes for the objects in the spherical image can be generated from the peaks, offsets and fields of views, and can be represented as $(\theta_i+\Delta\theta_i,\phi_i+\Delta\phi_i,\alpha_i,\beta_i)$.

\subsection{Implementation Details}\label{sec:impl_det}

\textbf{Ground Truth Generation.} We use the horizontal and vertical field of view of each object in the training dataset for the value of ground truth field of view $\mathbf{\hat{s}}_i$. To compute the ground truth offset $\mathbf{\hat{o}}_i$, we first transform the ground truth center point location from azimuthal and polar angle $(\hat{\theta}_i,\hat{\phi}_i)$ to 2D image coordinate $(\frac{\hat{\theta}_i WR}{2\pi},\frac{\hat{\phi}_i HR}{\pi})$ of the input image, where $WR\times HR$ is the resolution of the input image and $R$ is the downsampling factor. This location is mapped to the location $(\lfloor\frac{\hat{\theta}_i W}{2\pi}\rfloor,\lfloor\frac{\hat{\phi}_i H}{\pi}\rfloor)$ in the predicted heatmap due to downsampling. This new location corresponds to center point with azimuthal and polar angle $(\lfloor\frac{\hat{\theta}_i W}{2\pi}\rfloor\frac{2\pi}{W},\lfloor\frac{\hat{\phi}_i H}{\pi}\rfloor\frac{\pi}{H})$. Therefore, the ground truth offset is given as 
\begin{equation}
\mathbf{\hat{o}}_i=\left(\hat{\theta}_i-\left\lfloor\frac{\hat{\theta}_i W}{2\pi}\right\rfloor\frac{2\pi}{W},\hat{\phi}_i-\left\lfloor\frac{\hat{\phi}_i H}{\pi}\right\rfloor\frac{\pi}{H}\right).
\end{equation}
For the generation of ground truth heatmap, we also assign non-zero values to the negative locations within a radius of the positive location as in \cite{CornerNet} and \cite{CenterNet}. The radius is determined by ensuring that the locations within the radius would generate a bounding box with at least $t=0.7$ IoU with the ground-truth annotation. Given the radius, the value of ground truth heatmap is given by the function $\exp\left(-\frac{\arccos\left(\langle\mathcal{T}(\hat{\theta}_i,\hat{\phi}_i ),\mathcal{T}(\theta,\phi )\rangle\right)}{2\sigma^2}\right)$, where $(\hat{\theta}_i,\hat{\phi}_i )$ is the ground truth positive location, $(\theta,\phi)$ is the negative location within the radius, $\sigma$ is an adaptive standard deviation depending on the radius. As our unbiased spherical IoU is more complex than its planar counterpart, the calculation of radius is also more complex in the spherical case. We give the computation details in the supplementary material.

\vspace{0.1cm}
\noindent\textbf{Spherical Convolution.} As the spherical images suffer from distortion problem, we adopt spherical convolution in our network. Here we use tangent images \cite{TangentImage}, which is a spherical image representation that facilitates transferable and scalable $360^\circ$ computer vision. This representation renders a spherical image to a set of distortion-mitigated, locally-planar image grids tangent to a subdivided icosahedron. Standard CNNs can then be trained on these tangent images. The output feature maps can finally be rendered back to a sphere as the feature map of original spherical image. Thus in our Spherical CenterNet, the heatmap, offset and field of view are predicted for each tangent image, and then they are rendered back to the sphere for loss computation. We use this type of spherical convolution based on two considerations: it should keep the parameter sharing property of convolution; it does not lead to performance degradation if more convolutional layers are used. And we do not use other types of spherical convolutions, as they either break the parameter sharing property of convolution operation \cite{SphFeature,KTN} or would lead to performance degradation if more convolution layers are used in the network \cite{DistortionAware,SphereNet,DistortionECCV}.

\vspace{0.1cm}
\noindent\textbf{Training Details.} Our method is implemented in PyTorch \cite{pytorch} and 8 GeForce RTX 2080Ti GPUs are used for training with a batch size of 32 (4 images per GPU). The network is randomly initialized and trained from scratch. We use Adam \cite{Adam} to optimize the overall parameters objective for 160 epochs with the initial learning rate $1.25\times 10^{-4}$, and the learning rate is divided by 10 at 90 and 120 epochs. The input resolution of the whole network is $1024\times 512$, which develops into the output resolution of $256\times 128$ through the model. During training, because of the particularity of Equirectangular projection, we only use random flip as data augmentation without any other bells and whistles. For the training loss of 360-Indoor dataset, we set $\lambda_{off}=60$ and $\lambda_{fov}=10$ to balance the orders of magnitude for each loss term. And we find that $\lambda_{off}$ is acceptable in the range of 50 to 100, otherwise the poor performance is obtained. For that of the other two 360-VOC-Uniform and 360-VOC-Gaussian datasets, we 
keep $\lambda_{off}=1$ and $\lambda_{fov}=0.1$ in line with the original loss weights because each image only contains one object in these two datasets. See Section \ref{sec:data} for more details of the three datasets.

\section{Experimental Result}\label{sec:experiment}

In this section, we first compare our unbiased evaluation criteria with existing criteria and show that ours is more reasonable. Then, we compare our spherical image object detection algorithm with other state of the art methods. Finally, we give ablation studies.

\subsection{Criteria Comparison}\label{sec:CriteriaComparison}

We compare different evaluation criteria and show that our IoU calculation is unbiased through some toy examples. For each example, we randomly set the parameters of the ground truth and the predicted bounding box, and compute the IoU between them with different methods. For each case, we change the resolution of the underlying spherical images, so that the reference method based on spherical integral would be more precise. Please note that the resolutions set here is just for illustration purpose. We do not use such a high resolution for network training and inference.

From Table \ref{tab:Criteria} we can see that the IoUs computed with our method are more close to those computed with the reference method. It is no doubt that the first three methods give incorrect result, as they do not compute the IoU on the sphere. SphIoU \cite{AAAICriteria} also gives incorrect result, as it treats spherical rectangle as parts of spherical zones and has made too many approximations as introduced in Section \ref{sec:ExistingCriteria}. For the first two cases, the errors of SphIoU are even larger than the errors of IoUs computed based on polygon and circle representations. In addition, both our method and the penultimate spherical integral method give correct results, except that they are based on analytical solution and numerical integration respectively. This means that the accuracy of spherical integral method depends on the resolution of spherical images. As shown in Table \ref{tab:Criteria}, it is obvious that with the increase of the resolution, the accuracy of this method becomes higher gradually. Furthermore, this method is time consuming. It takes 37.5ms for IoU calculation, while our method only needs 0.99ms.

\begin{table}[!t]
	\caption{The IoUs computed with different methods for three cases. Here spherical integral is taken as the \textbf{reference method}, whose precision will be improved if underlying spherical images with high resolution are used.}
	\footnotesize
	\label{tab:Criteria}
	\centering
	\begin{tabular}{c|c|c|c|c}
		\toprule
		Cases&Methods&$8k\times 4k$&$10k\times 5k$&$12k\times 6k$\\
		\midrule
		\multirow{6}{*}{\includegraphics[width=0.3\linewidth]{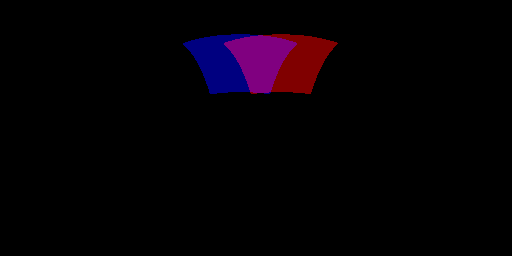}} &Rectangle \cite{360Indoor}&0.47115 & 0.47127&0.47163\\
		&Polygon \cite{SphereNet}&0.35911 & 0.35906 & 0.35891\\
		&Circle \cite{SpherePHD}&0.24278 & 0.24336 & 0.24286\\
		&SphIoU \cite{AAAICriteria}&0.16537 &0.16537 &0.16537 \\
		&Sph. Integral& \underline{0.32022} & \underline{0.32012} & \underline{0.32006}\\
		&Ours& \textbf{0.31974} & \textbf{0.31974} & \textbf{0.31974} \\
		\midrule
		\multirow{6}{*}{\includegraphics[width=0.3\linewidth]{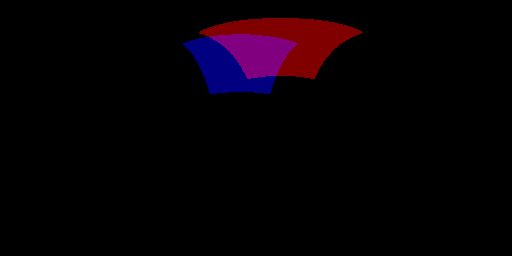}}&Rectangle \cite{360Indoor}& 0.55124&0.55120 &0.55155 \\
		&Polygon \cite{SphereNet}& 0.26954 & 0.26948 & 0.26958\\
		&Circle \cite{SpherePHD}& 0.24998 & 0.25061 & 0.24996\\
		&SphIoU \cite{AAAICriteria}& 0.11392& 0.11392 &0.11392\\
		&Sph. Integral& \underline{0.25816} & \underline{0.25807} & \underline{0.25801}\\
		&Ours& \textbf{0.25772} & \textbf{0.25772} & \textbf{0.25772}\\
		\midrule
		\multirow{6}{*}{\includegraphics[width=0.3\linewidth]{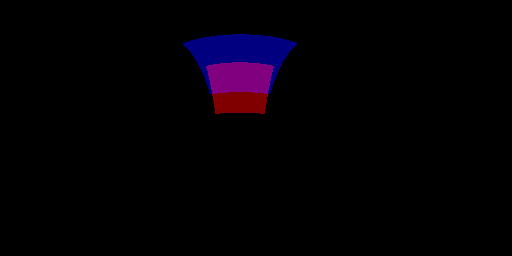}}&Rectangle \cite{360Indoor}&0.25930 &0.25873 & 0.25870\\
		&Polygon \cite{SphereNet}&0.31598 & 0.31542 & 0.31526\\
		&Circle \cite{SpherePHD}& 0.35973 & 0.35977 & 0.35992\\
		&SphIoU \cite{AAAICriteria}& 0.34220& 0.34220 &0.34220\\
		&Sph. Integral& \underline{0.33981} & \underline{0.33972} & \underline{0.33966}\\
		&Ours& \textbf{0.33935}&\textbf{0.33935} & \textbf{0.33935}\\
		\bottomrule
	\end{tabular}
\end{table}

\subsection{Comparison with Other Spherical Detectors}\label{sec:data}
\noindent\textbf{Dataset.} We conduct the experiments on three datasets, including one real-world dataset \textit{360-Indoor} \cite{360Indoor} composed of indoor $360^\circ$ spherical images for object detection, and another two synthetic spherical datasets \textit{360-VOC-Uniform} and \textit{360-VOC-Gaussian} generated from PASCAL VOC 2012 \cite{VOC15}.

\begin{figure}[!t]
	\centering
	\includegraphics[width=0.99\linewidth]{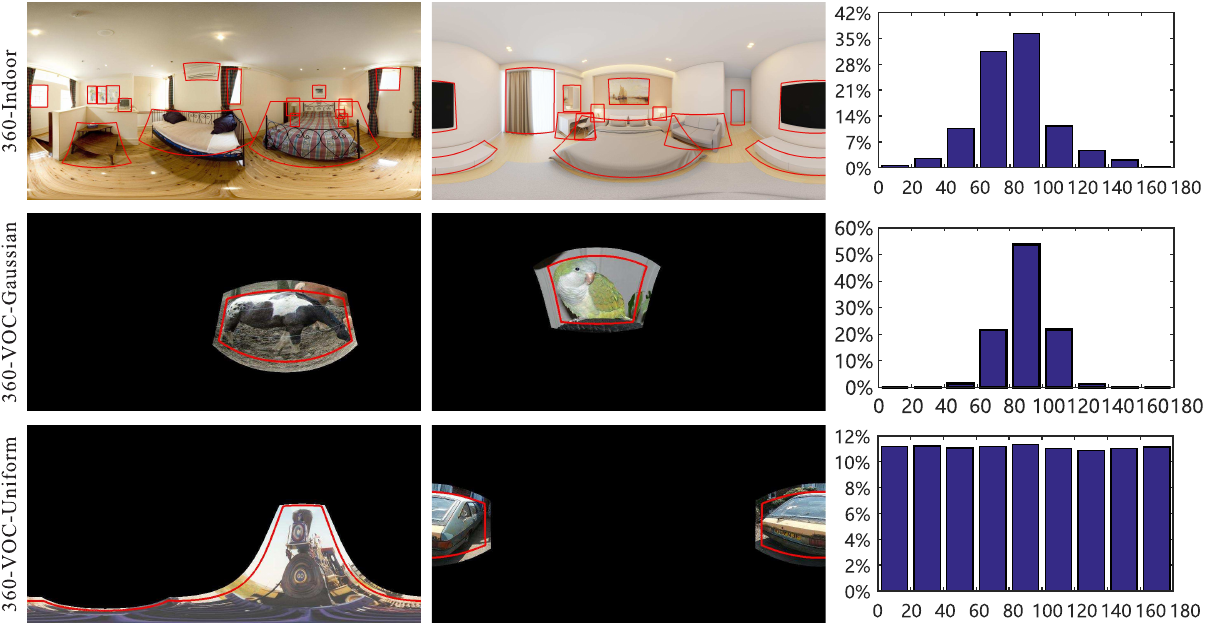}
	\caption{Some example images from 360-Indoor, 360-VOC-Gaussian and 360-VOC-Uniform datasets. Here we also plot the distribution of polar angles of the objects in each dataset.}
	\label{fig:dataset}
\end{figure}

\textit{360-Indoor} is a $360^\circ$ indoor dataset with 37 different categories specially designed for object detection task. This dataset has approximately 3k images and 90k labels in sum, which means that there are about 30 bounding boxes in each spherical images. In addition, this dataset uses spherical rectangle $(\theta,\phi,\alpha,\beta)$ for bounding box parameterization, instead of the commonly used planar rectangles. As shown in Figure \ref{fig:dataset}, most of the objects in this dataset are located near the equatorial region.

\textit{360-VOC-Gaussian} is a synthetic $360^\circ$ dataset generated from PASCAL VOC 2012 for object detection. It has 20 categories, and only one object instance is rendered in each spherical image. The objects in 360-VOC-Gaussian are normally distributed in the spherical images, and the size of the background is no less than half of the instance size if the instance is not at the edge of the image, otherwise the size of background is set as a random value. This dataset has a total of 28k images, including 18.6k images for training, 6.3k images for validating, and 3.1k images for testing, which follows the 6:2:1 splitting strategy for dataset approximately. Some example images are shown in Figure \ref{fig:dataset}. 

\textit{360-VOC-Uniform} is another synthetic $360^\circ$ dataset, and the only difference between 360-VOC-Uniform and 360-VOC-Gaussian, as shown in Figure \ref{fig:dataset}, is that the object instance is located at arbitrary position on the sphere with background from its original image rather than that normally distributed in 360-VOC-Uniform. Other parts, such as dataset source, categories, image sizes and so on, remain the same as 360-VOC-Uniform dataset.

\vspace{0.1cm}\noindent\textbf{Baseline Methods.} We compare our spherical CenterNet with three existing object detection algorithms for spherical images.


\textit{Multi-Kernel} \cite{DetectMultikernel} applies a multi-kernel layer after ROI Pooling layer in standard Faster R-CNN and incorporates position information of each proposal for object detection in spherical images.

\textit{Sphere-SSD} \cite{SphereNet} adapts SSD \cite{SSD} to spherical images and defines the anchor boxes based on the tangent planes of the sphere. 


\textit{Reprojection R-CNN} \cite{AAAICriteria} is a two-stage spherical object detector, where the first stage outputs spherical region proposals and the second stage refines the proposals predicted by the first stage.

As our network is based on the architecture of CenterNet, we also take the planar CenterNet \cite{CenterNet} as one of the baseline methods. To make these methods comparable, we set the networks of different methods to have the same backbone.

\begin{wrapfigure}{r}{0.65\textwidth}
  \begin{center}
    \includegraphics[width=\linewidth]{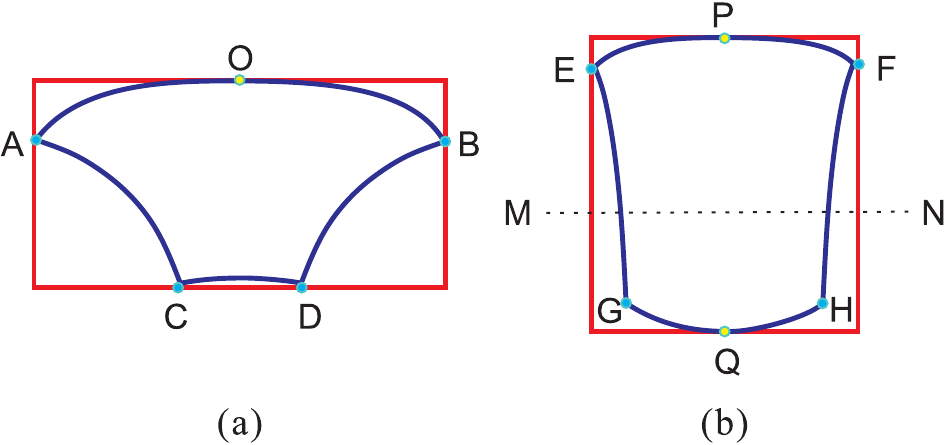}
	\caption{Two cases for converting rectangles to spherical bounding rectangles, (a) the  rectangle falling on the hemisphere, (b) the rectangle falling on the both sides of the  equator.}
	\label{fig:p2r}
  \end{center}
\end{wrapfigure}

\vspace{0.1cm}\noindent\textbf{Metric.} We use standard mAP \cite{VOC} in all categories as the evaluation metric for object detection. Please note that as the original evaluation metrics used in the baseline methods are biased, we transform the bounding boxes they predicted to spherical rectangles and use our unbiased IoU for the metric calculation.

It is also worth noting that we use rectangles tightly bounding spherical rectangles as the supervision for conventional methods and train networks to predict them. After these rectangles are predicted, we convert them to
spherical rectangles. The situation can be divided into two cases: the rectangle falling on the hemisphere
as shown in Figure \ref{fig:p2r} (a) and that falling on the both sides of the equator as shown in Figure \ref{fig:p2r} (b). Take for the first case (a) that the rectangle falling on the northern hemisphere as an example, and so is that falling on the southern hemisphere. The whole conversion process consists of three steps. We first determine the top side of the spherical rectangle (Corner O), and then we compute the top left and top right corners (Corner A and Corner B). Finally, the two bottom corners (Corner
C and Corner D) can be found by using the fact that the angle between them equals to the angle between
the two top corners. Given four computed corners, the spherical rectangle can be determined.



\begin{table*}[!t]
	\caption{The performance of different methods on 360-Indoor, 360-VOC-Uniform and 360-VOC-Gaussian datasets.}
	\small
	\label{tab:performance}
	\resizebox{\textwidth}{12mm}{
	\begin{tabular}{c|c|rrr|rrr|rrr}
		\toprule
		\multirow{2}{*}{Methods} & \multirow{2}{*}{Backbone} & \multicolumn{3}{c|}{360-Indoor} & \multicolumn{3}{|c|}{360-VOC-Gaussian} & \multicolumn{3}{c}{360-VOC-Uniform} \\
		
		& & $AP$ & $AP^{50}$ & $AP^{75}$ & $AP$ & $AP^{50}$ & $AP^{75}$ & $AP$ & $AP^{50}$ & $AP^{75}$\\
		\midrule
		CenterNet \cite{CenterNet}&ResNet-101	& 8.6 &	20.5 &	5.8  & 43.3 & 81.9 & 40.3 &	8.3 &	14.1 &	8.8\\
		Multi-Kernel \cite{DetectMultikernel}& ResNet-101 &	4.7 &	11.1 &	2.8 & 55.9 & 77.7 & 64.8 & 7.0 &	12.5 &	7.3 \\
		Sphere-SSD \cite{SphereNet}&ResNet-101	& 2.9 &	7.8 &	1.4 & 21.8 & 28.4 & 26.7 &	11.7 &	19.2 &	13.4 \\
		Reprojection R-CNN \cite{AAAICriteria}&ResNet-101&	5.0 &	15.3 &	1.9 & 53.6 & 62.2 & 44.8 &	9.5 &	13.8 &	10.1 \\
		Ours  & ResNet-101 & \textbf{10.0} &	\textbf{24.8} &	\textbf{6.0} & \textbf{65.5} & \textbf{84.6} & \textbf{75.5} &	\textbf{15.8} &	\textbf{21.5} &	\textbf{18.1} \\
		\bottomrule
	\end{tabular}}
\end{table*}

\vspace{0.1cm}\noindent\textbf{Quantitative Results.} The performance of different methods on the three dataset are shown in Table \ref{tab:performance}. For each dataset, we give the $AP$, $AP^{50}$ and $AP^{75}$ performance. From the table, we can see that our method can give the best performance on all the three datasets. Compared with the performance on 360-Indoor, our method gives lower $AP^{50}$ on 360-VOC-Uniform. The representation we used for spherical convolution involves projecting each spherical image onto 20 tangent images. Thus the object in each image of 360-VOC-Uniform may be split into two or more different tangent images, which would degrade the detection performance. Reprojection R-CNN also gives lower $AP^{50}$ on 360-VOC-Uniform. This is because the method is trained with biased IoU calculation. If the predicted or ground truth bounding boxes are located near the polar regions, e.g. the characteristic of 360-VOC-Uniform dataset, the IoU calculation of Reprojection R-CNN would give large errors as we discussed in Section \ref{sec:CriteriaComparison}. 

Meanwhile, because the objects in 360-Indoor and 360-VOC-Gaussian datasets are less distorted than those in 360-VOC-Uniform, planar CenterNet performs better on 360-Indoor. Unlike these methods mentioned before, Sphere-SSD gives better performance on 360-VOC-Uniform. The reason is that Sphere-SSD adopts spherical convolutions to deal with image distortions in its network, which is basically consistent with the projection mode of the synthetic dataset so that more objects in polar regions can be detected. As to 360-Indoor and 360-VOC-Gaussian datasets, the disadvantages of Sphere-SSD are exposed as seen from the detection results.

When synthesizing 360-VOC-Uniform dataset, the polar angles of rendered objects follow a uniform distribution. This is different with the strategy of 360-Indoor and would lead to lower performance. Thus we synthesize another dataset called 360-VOC-Gaussian, in which the polar angles of objects follow a standard gaussian distribution as 360-Indoor dataset. For this dataset, we conduct each method on it and then get the conclusion that the detection results of these methods have similar trends with those of 360-Indoor, which corroborates our previous results analysis. Moreover, Multi-Kernel gives better performance on 360-VOC-Gaussian. The main reason is mainly that each image in 360-VOC-Gaussian only contains one object with less distortion.

In addition, the other three baselines have worse performances than CenterNet, and the reason of which is quite likely to be that they are all anchor-based methods, while CenterNet is anchor-free method. As shown in previous work \cite{zhang2020bridging}, the performance of anchor-based method is more easily affected by how to select positive and negative training samples. As they use biased IoU calculation, positive and negative training samples may be selected incorrectly and this leads to poor performance.

\begin{figure*}[!t]
	\centering
	\includegraphics[width=\linewidth]{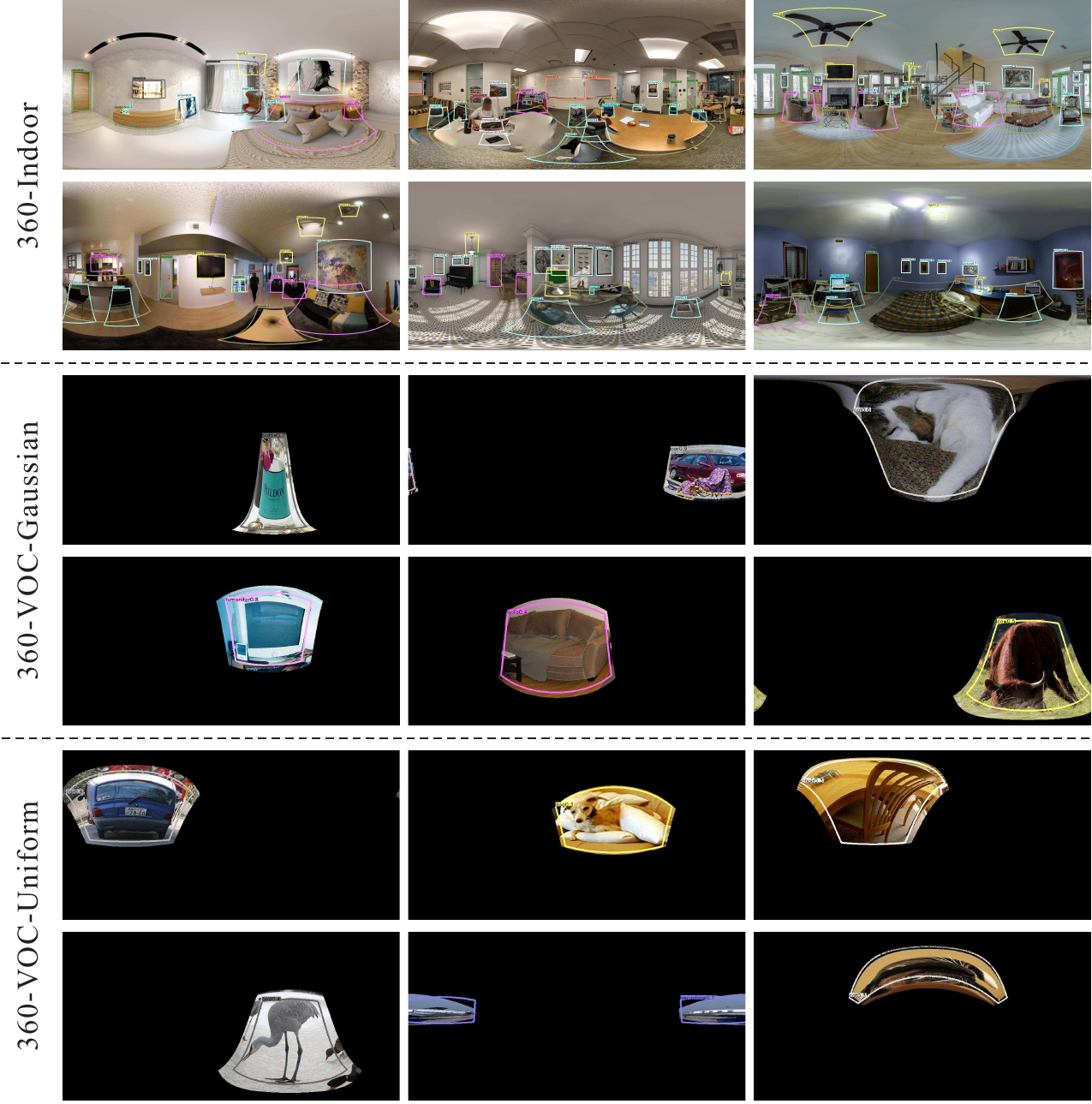}
	\caption{Visual detection results of our method on 360-Indoor, 360-VOC-Gaussian and 360-VOC-Uniform datasets.}
	\label{fig:result}
\end{figure*}

\vspace{0.1cm}\noindent\textbf{Visual Detection Results.} We give some visual detection results of our method on the three datasets in Figure \ref{fig:result}. Our method can successfully detect the objects in spherical images, even if the objects have large distortions or are split by the left or right boundaries of spherical images. See the aeroplane and the cat in images from 360-VOC-Uniform, the table and the bed in images from 360-Indoor, and the cat and the bottle in images from 360-VOC-Gaussian.

\begin{figure}[!t]
	\centering
	\includegraphics[width=\linewidth]{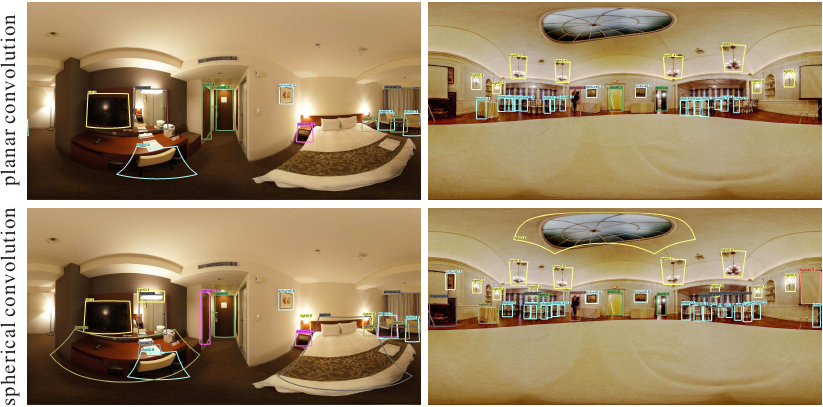}
	\caption{Compared with planar convolution, spherical convolution can detect more seriously distorted objects.}
	\label{fig:Ablation}
\end{figure}

\subsection{Ablation Study}

\begin{table}[!b]
	\caption{The performance of our network with different backbones and different types of convolutions.}
	\label{tab:Ablation}
	\centering
	\begin{tabular}{c|c|rrr}
		\toprule
		Backbone & Convolution& $AP$ & $AP^{50}$ & $AP^{75}$\\
		\midrule
		ResNet-101 & spherical & 10.0 &	24.8 &	6.0 \\
		Hourglass & spherical & \textbf{14.1} &	\textbf{31.4} &	\textbf{11.0} \\
		\midrule
		Hourglass & planar &	12.7 &	28.4 &	9.3 \\
		Hourglass & spherical & \textbf{14.1} &\textbf{	31.4} &	\textbf{11.0} \\
		\bottomrule
	\end{tabular}
	
\end{table}

To more throughly analyze our method, we carry out two types of ablation studies.

\vspace{0.1cm}\noindent\textbf{Backbone.} We train our network with two different backbones: ResNet-101 \cite{ResNet} and Hourglass \cite{Hourglass}. These two backbones have about the same depth, but Hourglass uses skip layers to bring back the details to the upsampled features. This can greatly improve the performance of the detection network, as shown in Table \ref{tab:Ablation}.

\vspace{0.1cm}\noindent\textbf{Type of Convolution.} Our network leverages spherical convolutions to deal with the distortions of $360^\circ$ spherical images. To check the effect of spherical convolutions, we have also trained a network using traditional planar convolutions and compared it with the network using spherical convolutions. As shown in Table \ref{tab:Ablation}, the usage of spherical convolutions can significantly improve the detection performance. We give some visual comparisons in Figure \ref{fig:Ablation}. We can see that spherical convolutions can let the network to detect the objects with large distortions. For example, the seriously stretched table, bed and light are detected by the network using spherical convolutions.

\section{Conclusion}\label{sec:conclusion}

Spherical object detection will become more and more important along with the fact that $360^\circ$ spherical images can be easily captured nowadays. In this paper, we propose the first unbiased IoU for spherical image object detection. We first illustrate that spherical rectangles are natural representations for the bounding boxes of spherical objects, then we give the unbiased IoU calculation method based on the new representations. Compared with our proposed metric, existing evaluation metrics are all biased. We also present a new anchor-free object detection algorithm for spherical images, which directly output bounding boxes for objects. Extensive experiments on three datasets show that our method can get better results. In the future, we would like to apply our unbiased IoU to other types of spherical image processing or in other computer vision tasks like visual tracking.

\bibliography{iclr2021_conference}
\bibliographystyle{iclr2021_conference}

\appendix
\section{Appendix}
In this appendix (supplemental material), we introduce the derivation of the formula for the area of spherical rectangle given its field of view $\alpha$ and $\beta$. We also describe the details of how to compute the radius used to generate the ground truth heatmap.

\subsection{Area of Spherical Rectangle}

\begin{figure}[!h]
	\centering
	\includegraphics[width=0.8\linewidth]{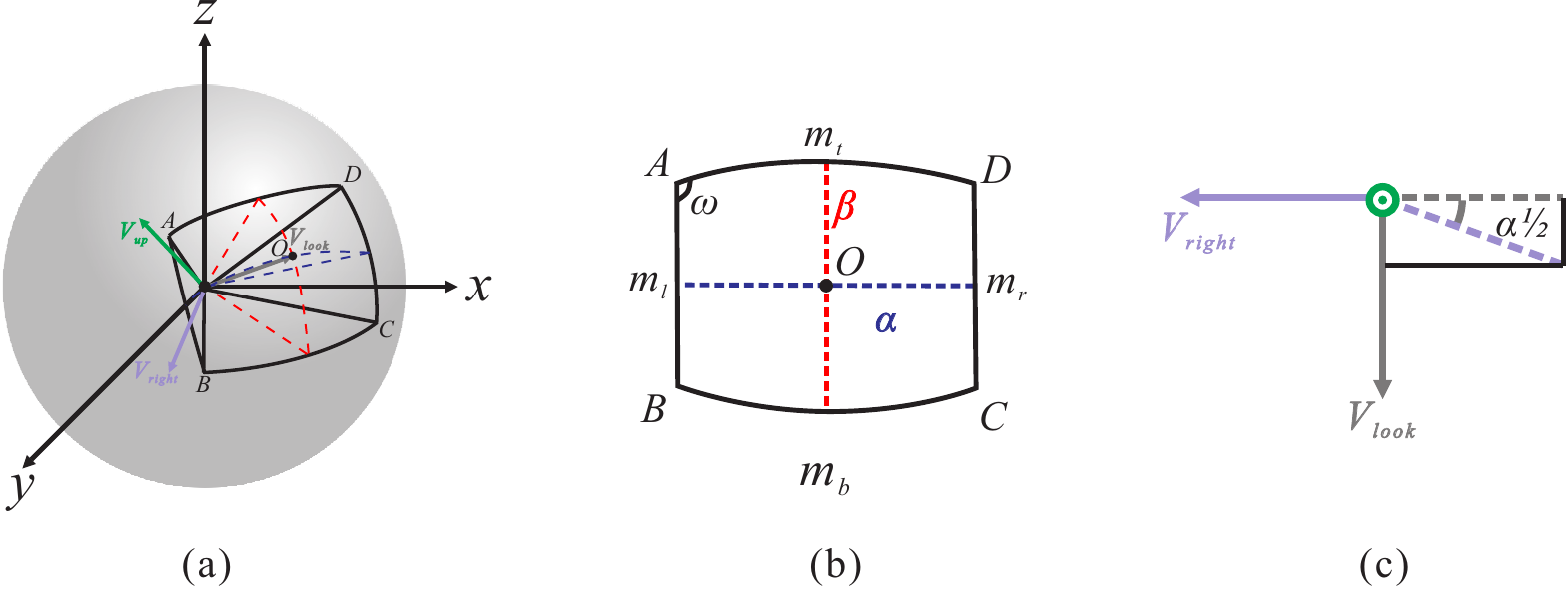}
	\caption{(a) To compute the area of a spherical rectangle, we establish a local coordinate system $V_{look}V_{right}V_{up}$. (b) Then the planes corresponding to four sides of the spherical rectangle can be obtained by rotating the planes determined by the center point of the spherical rectangle. (c) The normals of the planes can be easily computed based on the established coordinate system, which can be used to compute the area.}
	\label{fig:area}
\end{figure}

According to the definition of spherical polygons, we know that the four sides of each spherical rectangle are all great-circle arcs, which are the intersection of the surface with planes through the center of the sphere \cite{enwiki:1016967508}. We also know that the four angles $\{\omega_i,i=1,\cdots,4\}$ of each spherical rectangle are equal due to symmetry. Therefore we can compute the area of a spherical rectangle $b$ as $A(b)=4\omega-2\pi$ according to the formula for the area of spherical polygons, where $\omega$ is the angle of spherical rectangle and is defined as the angle between the planes that the neighboring sides of each spherical rectangle lie on. Thus, the key step is how to compute the value of angle $\omega$.

The angle between two planes equals to the angle between the normals of these two planes. For each spherical rectangle, the four planes corresponding to its four sides can be obtained by rotating the planes that pass through the center point $(\theta,\phi)$ of the spherical rectangle. Taking the left plane $P_{Am_lB}$ as an example, as shown in Figure \ref{fig:area} (a) and (b), it can be obtained by rotating the plane $P_{m_tOm_b}$ around the axis $V_{up}$ by $\frac{\alpha}{2}$, where $\alpha$ is the horizontal field of view of the spherical rectangle. The $V_{up}$ is an axis of the coordinate system that we established based on axis $\vec{z}=[0,0,1]^T$ and the azimuthal and polar angle $(\theta,\phi)$ of the center point of the spherical rectangle
\begin{equation}
\begin{cases}
V_{look}=\left[ \sin(\phi)\cos(\theta),\sin(\phi)\sin(\theta),\cos(\phi)\right]^T \\
V_{right}=[-\sin(\theta),\cos(\theta),0]^T\\
V_{up}=[-\cos(\phi)\cos(\theta),-\cos(\phi)\sin(\theta),\sin(\phi)]^T
\end{cases}
\end{equation}
As a consequence, the normal of the left plane $P_{Am_lB}$ can also be obtained by rotating the normal of plane $P_{m_tOm_b}$ around the axis $V_{up}$ by $\frac{\alpha}{2}$. This is illustrated in Figure \ref{fig:area} (c), in which the dashed gray line and the dashed blue line denote the normal of the plane $P_{m_tOm_b}$ and $P_{Am_lB}$ respectively. Thus, the normal of the left plane $P_{Am_lB}$ can be derived as
\begin{equation}
N_l=\sin\frac{\alpha}{2}V_{look}-\cos\frac{\alpha}{2}V_{right}.
\end{equation}
Similarly, we can also get the normal of the top plane as 
\begin{equation}
N_t=\sin\frac{\beta}{2}V_{look}-\cos\frac{\beta}{2}V_{up}.
\end{equation}
The angle $\omega$ is given by
\begin{equation}\label{equ:dot_pro}
\omega=\pi-\arccos(\langle N_l,N_t\rangle)=\arccos(-\sin\frac{\alpha}{2}\sin\frac{\beta}{2})
\end{equation}
where $\langle\cdot,\cdot\rangle$ is the dot product of two input vectors.

The normal of the right plane $N_r$ and the normal of the bottom plane $N_b$ is given in the similar way
\begin{equation}
N_r=\sin\frac{\alpha}{2}V_{look}+\cos\frac{\alpha}{2}V_{right}.
\end{equation}
\begin{equation}
N_b=\sin\frac{\beta}{2}V_{look}+\cos\frac{\beta}{2}V_{up}.
\end{equation}
Then we can calculate the four angles through Equation \ref{equ:dot_pro} and most interestingly find that they are all equal.

After obtaining the four angles, the area $A(b)$ of spherical rectangle $b$ can be consequently computed as
\begin{equation}
A(b)=4\arccos(-\sin\frac{\alpha}{2}\sin\frac{\beta}{2})-2\pi
\end{equation}

\subsection{The Calculation of Radius}
\begin{figure}[!h]
	\centering
	\includegraphics[width=0.8\linewidth]{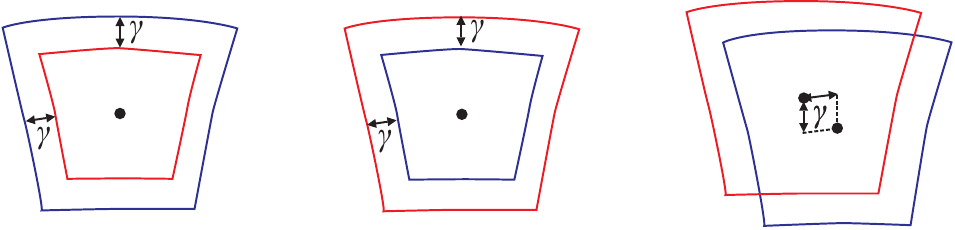}
	\caption{Different relationships between the ground truth (red) and the predicted (blue) bounding boxes: (a) the predicted bounding box contains the ground truth bounding box, (b) the ground truth bounding box contains the predicted bounding box, (c) these two bounding boxes intersect with each other.}
	\label{fig:radius}
\end{figure}

When generating the ground truth heatmap, we assign non-zero values to the negative locations within a radius of the positive location as in Corner \cite{CornerNet} and CenterNet \cite{CenterNet}. Here we describe how to calculate the radius. Actually, there are three cases for the radius, which correspond to different relationships between ground truth and predicted bounding boxes as shown in Figure \ref{fig:radius}. We only derive for the first case in detail, and the other two cases can be derived in a similar way as follows.

\textbf{Case a:}
Let $\alpha$ and $\beta$ be the field of view of ground truth bounding box, and let $\gamma$ be the radius. The IoU, whose threshold is $t$, between the predicted and the ground truth bounding box is
\begin{equation}\label{equ:threshold}
\frac{4\arccos(-\sin\frac{\alpha}{2}\sin\frac{\beta}{2})-2\pi}{4\arccos\left(-\sin(\frac{\alpha+2\gamma}{2})\sin(\frac{\beta+2\gamma}{2})\right)-2\pi}=t.
\end{equation}
According to the product-to-sum identities of trigonometric functions,
\begin{equation}
\sin(\frac{\alpha+2\gamma}{2})\sin(\frac{\beta+2\gamma}{2})=-\frac{1}{2}\left[\cos(\frac{\alpha+\beta+4\gamma}{2})-\cos(\frac{\alpha-\beta}{2})\right].
\end{equation}
Thus Equation \ref{equ:threshold} can be written as
\begin{equation}
2\arccos(-\sin\frac{\alpha}{2}\sin\frac{\beta}{2})-\pi=t\left[ 2\arccos\left(\frac{1}{2}\left[\cos(\frac{\alpha+\beta+4\gamma}{2})-\cos(\frac{\alpha-\beta}{2})\right]\right)-\pi\right]. 
\end{equation}
The above equation can be reduced to
\begin{equation}
\cos(\frac{\alpha+\beta+4\gamma}{2})=2\cos\left(\frac{\arccos(-\sin\frac{\alpha}{2}\sin\frac{\beta}{2})-\frac{\pi}{2}}{t}+\frac{\pi}{2}\right)+\cos(\frac{\alpha-\beta}{2}),
\end{equation}
then
\begin{equation}
\gamma = \frac{1}{2}\arccos\left(-2\sin\left(\frac{\arcsin(\sin\frac{\alpha}{2}\sin\frac{\beta}{2})}{t}\right)+\cos(\frac{\alpha-\beta}{2})\right)-\frac{\alpha+\beta}{4}.
\end{equation}
The derivation of \textbf{Case b} and \textbf{Case c} is the same as above, thus we only give the final radius calculating formulas. 

\textbf{Case b:}
\begin{equation}
\gamma = -\frac{1}{2}\arccos\left(-2\sin\left({t}\arcsin(\sin\frac{\alpha}{2}\sin\frac{\beta}{2})\right)+\cos(\frac{\alpha-\beta}{2})\right)+\frac{\alpha+\beta}{4}.
\end{equation}

\textbf{Case c:}
\begin{equation}
\gamma = -\arccos\left(-2\sin\left(\frac{{2t}
\left(\arccos(-\sin\frac{\alpha}{2}\sin\frac{\beta}{2})-2\pi\right)}{1+t}\right)+\cos(\frac{\alpha-\beta}{2})\right)+\frac{\alpha+\beta}{2}.
\end{equation}
The final radius is the minimum of the above three cases, which is
\begin{equation}
\gamma = \min(\gamma\bm{_a}, \gamma\bm{_b}, \gamma\bm{_c})
\end{equation}
where $\gamma\bm{_a}, \gamma\bm{_b}, \gamma\bm{_c}$ represent the radius of the above three cases respectively. 

\end{document}